\documentclass[10pt,conference]{IEEEtran}
\IEEEoverridecommandlockouts
\usepackage{algorithm,algpseudocode} 
\usepackage{cite}
\usepackage[utf8x]{inputenc} 
\usepackage{amsmath}
\usepackage{amssymb,amsmath}
\usepackage{graphicx}
\usepackage{epsfig}
\usepackage{grffile}
\usepackage{balance,color}
\usepackage{caption}
\usepackage{subcaption}
\usepackage{xcolor, soul}
\sethlcolor{cyan}
\providecommand{\keywords}[1]{{\textit{Index Terms}}}

\begin{document}

\title{ PDSR: Efficient UAV Deployment for Swift and Accurate Post-Disaster Search and Rescue} 
\author{ Alaa Awad Abdellatif, Ali Elmancy, Amr Mohamed, Ahmed Massoud, Wadha Lebda,  Khalid K. Naji \\
\begin{tabular}{c}
	College of Engineering, Qatar University, Qatar \\ 
		Email: alaa.abdellatif@ieee.org,  \{ali.elmancy, amrm, ahmed.massoud, wadha.lebda, knaji\}@qu.edu.qa
		\thanks { This work was made possible by ARG grant \# ARG01-0527-230356 from the Qatar National Research Fund (a member of Qatar Foundation).  The findings achieved herein are solely the responsibility of the authors.  }  
\end{tabular}
}
\maketitle

\begin{abstract}
This paper introduces a comprehensive framework for Post-Disaster Search and Rescue (PDSR), aiming to optimize search and rescue operations leveraging Unmanned Aerial Vehicles (UAVs). The primary goal is to improve the precision and availability of sensing capabilities, particularly in various catastrophic scenarios. Central to this concept is the rapid deployment of UAV swarms equipped with diverse sensing, communication, and intelligence capabilities, functioning as an integrated system that incorporates multiple technologies and approaches for efficient detection of individuals buried beneath rubble or debris following a disaster. Within this framework, we propose architectural solution and address associated challenges to ensure optimal performance in real-world disaster scenarios. The proposed framework aims to achieve complete coverage of damaged areas significantly faster than traditional methods using a multi-tier swarm architecture. Furthermore, integrating multi-modal sensing data with machine learning for data fusion could enhance detection accuracy, ensuring precise identification of survivors. 

\end{abstract}

\section{Introduction\label{sec:Introduction}} 

Disaster management, a field involving preparation, response, and relief efforts in catastrophic events like earthquakes, post-war scenarios, and tsunamis, has gained significant attention over the past decade.  
Disaster management activities are generally divided into pre-disaster and post-disaster operations. Pre-disaster activities include prevention, detection, and preparedness phases, all aimed at minimizing risks and informing the public about potential hazards to reduce the impact of natural disasters. On the other hand, post-disaster activities include damage assessment, victim rescue, and efforts for repair and recovery. This paper focuses on enhancing Post-Disaster Search and Rescue (PDSR) operations through the use of Unmanned Aerial Vehicles (UAVs), presenting a novel approach that highlights their topological and sensing capabilities.  

UAVs, as multi-faceted Internet of things (IoT) devices equipped with onboard sensors, can play a pivotal role in disaster management \cite{9712460}. Such IoT devices, which may include cameras, microphones, and ultra-wideband (UWB) impulse radar, enable UAVs to collect critical data from disaster-affected areas. This data includes hazard maps, detailed building renderings, and comprehensive elevation models. With recent advancements in remote sensing technology, processing, and communication capabilities, UAVs are revolutionizing the future of disaster management by enabling the detection of victims trapped beneath debris.  
However, deploying these technologies effectively in real-world scenarios presents several challenges. The most significant issues are: i) the detection of human vital signs, which are often weak and easily masked by background noise and surrounding clutter, and ii) the limited sensing area that a single UAV can cover. 

Recent signal processing techniques have been proposed to address the challenges of sensing behind obstacles. For instance, the authors in \cite{tivive2020toward} introduced a target detection method for stepped-frequency through-the-wall radar that eliminates wall clutter while localizing both stationary and moving targets. Meanwhile, the authors in \cite{zheng2022recovering} presented a deep learning-based method for human pose and shape recovery from through-the-wall radar imaging (TWRI). This approach employs a teacher-student learning framework, where an optical system connected to the radar captures paired RGB and TWRI images. Computer vision techniques function as the teacher network, offering cross-modal supervision signals, thereby eliminating the need for manual labeling.  However, these techniques were primarily developed for static platforms and require data with relatively long-duration signals, making them unsuitable for drone applications.  


In this work, we propose a PDSR framework that utilizes a swarm of drones equipped with advanced sensing technology to locate survivors after catastrophic events. This framework aims to minimize operational time, ensure comprehensive and rapid coverage of affected areas, and achieve accurate and swift detection of trapped survivors. Within our framework, UAVs play a crucial role in sensing the environment to avoid obstacles, localize targets, and detect trapped survivors. Also, they can provide instant communication for users while operating in dual-functional radar-communication mode. However, the specific communication techniques used for disaster assessment are beyond the scope of this article.  
Thus, the contributions of this work are:  
\begin{enumerate}
    \item Disaster assessment is discussed from a novel architectural perspective. This includes exploring a UAV swarm architecture that offers a robust solution for detecting victims trapped under rubble while minimizing scanning and detection time, thereby increasing survival rates.  
	\item Proposing a swift, multi-modal solution based on UAV and multiple sensor technologies. This innovative approach integrates optical and radar data to improve the precision and efficiency of detecting human targets behind obstacles. 
	\item Implementing and demonstrating the effectiveness of the proposed framework in enhancing detection accuracy and latency, along with providing a comprehensive discussion of open problems and conceptual solutions for PDSR operations.  
\end{enumerate}

The following sections cover the PDSR challenges and related work (Section \ref{sec:Sec2}), the proposed PDSR architecture and its benefits (Section \ref{sec:system}), the design, implementation, and discussion of the proposed sensing solution (Section \ref{sec:Benefits}), and potential directions for future research (Section \ref{sec:conclusion}).

\section{Post-Disaster Search and Rescue \label{sec:Sec2}}

In this section, we highlight the major challenges associated with PDSR operations, considering insights from recent relevant literature.  
 
\subsection{Challenges of PDSR Operations \label{sec:why}} 

PDSR operations encompass a range of challenges in sensing, communications, and computing, all of which are critical for immediate response efforts, such as damage assessment and life-saving measures. Innovative UAV-based solutions have the potential to significantly enhance these operations by utilizing advanced remote sensing, computer vision, and communication capabilities to identify key aspects of disaster areas, including hazard mapping, locating victims trapped under rubble, and generating detailed building renderings. However, to fully harness the potential of UAVs in PDSR, several key challenges must be effectively addressed.

\textbf{Topological features of the disaster area: } 
Accurately determining the scale of the disaster-affected area, using vision-based analysis, is crucial for effective emergency response and resource allocation, especially in PDSR operations. It allows authorities to estimate the severity of the situation and strategically deploy UAVs for reconnaissance and rescue missions. Understanding the extent of the disaster enables precise calculations regarding the number and types of UAVs required to cover the area, ensuring optimal use of resources. UAVs equipped with various IoT sensors and imaging technologies can then gather critical data, such as high-resolution imagery, hazard maps, and information on victims trapped under debris. This knowledge empowers Search and Rescue (SAR) teams to optimize their efforts, enhancing efficiency and effectiveness in mitigating the disaster's impact and saving lives.


\textbf{Obstacles and rubble:}  
In PDSR operations, navigating obstacles and rubble is a significant challenge that hinders access to potential victims. These obstacles, including collapsed buildings and debris, often impede SAR teams from reaching survivors. The timely and accurate acquisition of information is critical in these scenarios, as it can mean the difference between life and death. However, traditional data collection methods during emergencies are often hampered by coordination issues among various agencies, coupled with delays in deploying SAR teams due to damaged infrastructure, unstable ground, and blocked roads. 

\textbf{Efficient under-rubble victims detection:} 
Detecting victims trapped under rubble in PDSR operations is especially challenging due to the complex environment and limited visibility available to cameras and rescue personnel. Overcoming this challenge requires the development of reliable and accurate detection systems that leverage radar sensors capable of penetrating debris to locate survivors. Critical considerations include selecting the most suitable radar technique, assessing the effectiveness of different IoT sensors in penetrating various materials, and integrating machine learning methods to improve the accuracy and efficiency of survivor detection.   

\textbf{ State of the network Infrastructure:} 
Before deploying UAVs for PDSR operations, it is crucial to evaluate the state of existing infrastructure. Some base stations (BSs) in the disaster zone may still be operational while others are not. Assessing the locations and conditions of these active BSs is essential for optimizing the type, quantity, and placement of UAVs. Collaborating with operational BSs can streamline PDSR efforts, ensuring efficient resource utilization and improving the overall effectiveness of disaster response.

\subsection{Related Work  \label{sec:Related}} 

  
During disasters, terrestrial communication infrastructure often faces heavy loads and is prone to disruptions, sometimes leading to partial or complete failure. This highlights the urgent need for a reliable backup network to maintain uninterrupted communication and coverage. A robust backup network is vital for enabling communication among emergency responders, affected communities, and relief organizations. Consequently, many researchers have explored the use of UAVs to support emergency services.  
For instance, UAVs can establish ad-hoc communication networks, serve as aerial communication relays, and provide temporary connectivity in areas where terrestrial communication infrastructure has been damaged or disrupted \cite{matracia2021topological, 10517504}.   
In \cite{sobot2023two}, the authors design a two-tier Low Power Wide Area Network (LPWAN) system using UAV base stations for dynamic deployment in deep rural environments. This UAV-based LPWAN network enhances the existing macro-cellular LPWAN network (Tier 1) by adding a second layer of mobile base stations (Tier 2), which extend connectivity to static or mobile LPWAN user equipment in areas without direct Tier 1 coverage.  
In \cite{mohammadisarab2024resilient}, the authors focus on enhancing connectivity in hierarchical non-terrestrial networks for disaster relief services by optimizing radio resources, computing capacities, and UAV trajectories. In this setup, UAVs serve as aerial base stations, linking low-capability users via power domain non-orthogonal multiple access (PD-NOMA).   
However, technical limitations persist, including the lack of widespread Low-Earth Orbit (LEO) satellite constellations or elevated terrestrial base station antennas for aerial backhaul. Moreover, insufficient data rates hinder the safe coordination of large UAV swarms. Overcoming these challenges and reducing deployment costs are critical areas for advancing UAV-enabled emergency communication systems.  

Joint communication and radar sensing (JCAS), also known as dual-functional radar-communication (DFRC), has gained significant interest recently \cite{chaccour2024joint}.  
In DFRC, a single transmitted signal is utilized, with most hardware and signal processing shared between communication and radar functions, thereby minimizing payload and resource usage. However, this approach may not be suitable for PDSR system due to the need for communication between  UAVs at High-Altitude (HA) while simultaneously sensing behind obstacles at Low-Altitude (LA).  
Other work has explored the integration of light-fidelity (LiFi) systems on low-altitude balloons for conducting SAR operations \cite{surampudi2018can}. This innovative approach demonstrates that LiFi-equipped balloons can serve as a cost-effective and highly versatile alternative to traditional radio frequency systems. What sets LiFi technologies apart is their ability to operate in diverse environmental conditions, including floods or areas with combustible gas leakages. The work in \cite{lai2024energy} presents an energy-efficient UAV-assisted hierarchical architecture for Wireless Sensor Networks (WSNs) designed to monitor remote environments. This architecture integrates terrestrial WSNs, a UAV, LEO satellites, and a cloud server, outperforming similar systems in terms of network average energy consumption, standard deviation of energy usage, network lifetime, and UAV energy efficiency.  
Through-wall human localization has also gained significant attention recently. Many studies employ cameras, Wi-Fi, or millimeter-wave radars for fine-grained localization, yet wall occlusion remains a major challenge. While Wi-Fi can penetrate walls, it struggles with precise localization due to its narrow bandwidth. Specialized sensors like thermal imaging cameras and acoustic sensors, mounted on UAVs or deployed on the ground, are also used \cite{10517950}. However, these sensors are sensitive to temperature variations and non-line-of-sight conditions, making it difficult to detect buried victims when temperature differences are minimal or other heat sources are present. 

Advanced imaging technologies such as ground-penetrating radar (GPR), LiDAR, and multiple-input–multiple-output (MIMO) radar are employed to capture localization, behavior information, and radar reflection heatmaps of individuals behind walls, or to create detailed 3D maps of disaster sites to identify voids and potential victim locations \cite{10121792, rohman2021through}. However, these methods have not addressed the integration of multi-modal data from UAVs for detecting and locating victims under rubble, which is crucial for effective PDSR operations. Thus, this paper is the first to tackle the challenges of detecting and rescuing victims trapped under debris by leveraging swift UAV deployment and multi-modal data collection and processing capabilities.   
\section{PDSR Architecture and  Benefits \label{sec:system}}

In this section, we outline the PDSR system architecture and highlight the key benefits of leveraging this architecture to optimize disaster response efforts.

\subsection{PDSR Architecture\label{sec:Architecture}} 

\begin{figure*}[]
	\centering
		\scalebox{2.2}{\includegraphics[width=0.33 \textwidth]{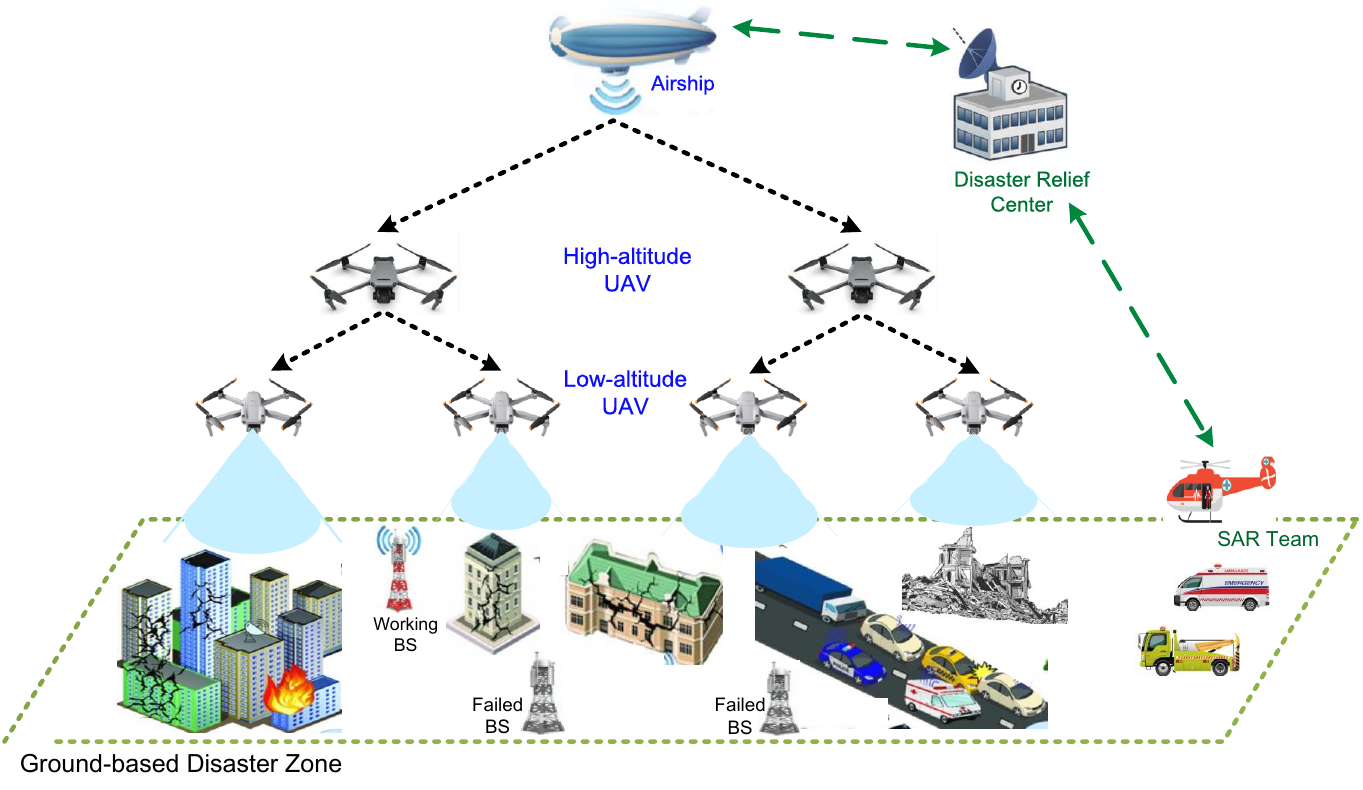}}
	\caption{The proposed PDSR system architecture. }
	\label{fig:system_model}
\end{figure*}

The considered architecture, as depicted in 
Figure~\ref{fig:system_model}, represents a comprehensive system model that is capable of conducting aerial surveys of disaster-affected areas. Thus, it includes the following basic components: 
\subsubsection{Disaster Relief center (DRC)}
The DRC serves as a centralized entity that manages and maintains full control over the disaster area, enabling rapid damage assessment, hazard identification, and evaluation of access routes for rescue operations. Specifically, the DRC is responsible for: (i) issuing initial control commands and assigning GPS coordinates to each UAV; (ii) storing mission data for future analysis and historical record-keeping; (iii) providing real-time information to the SAR team; and (iv) maintaining continuous updates on the status of each UAV in operation.  

\subsubsection{Airship}
Several innovative projects have been initiated to develop airships designed as advanced platforms for safety management  \cite{matracia2021topological}. In our architecture, these airships are utilized to transport multiple drones and deploy them near disaster zones, enabling rapid response efforts. Moreover, the airships can establish video communication links between civilians and first responders, enhancing coordination and support during emergencies. Equipped with 5G or LTE technologies, these airships can also connect seamlessly with the DRC to enable command, control, and communication functions. 

\subsubsection{High-altitude UAVs (HA-UAVs)} 
High-altitude UAVs possess the capability to achieve higher autonomy and signal strength owing to their restricted mobility and reduced need for frequent relocation. This limitation in mobility can actually confer certain advantages, particularly in adverse meteorological conditions, where the risk of collision is significantly diminished. By operating at elevated altitudes and maintaining relatively stationary positions, these UAVs can ensure more stable communication links and perform extended-duration missions with enhanced reliability. Moreover, their reduced susceptibility to atmospheric disturbances allows them to maintain consistent performance, making them well-suited for deployment in challenging environments or during inclement weather conditions. Thus, while HA-UAVs may exhibit limited mobility, they offer distinct advantages in terms of operational robustness and performance stability, particularly in adverse or unpredictable conditions.

\subsubsection{Low-altitude UAVs (LA-UAVs)}
These drones autonomously proceed to pre-defined locations designated by HA-UAVs or the DRC, where they carry out detection tasks before returning to the airship/DRC. Their primary focus is on under-rubble detection, aiming to identify humans trapped beneath debris using a variety of sensing techniques integrated into the LA-UAVs. These drones are also equipped with data fusion capabilities, allowing them to combine detection results from different IoT devices  to accurately determine the presence of a human. The gathered data, along with precise location coordinates, is then communicated to HA-UAVs and the DRC, facilitating quicker response times for rescue teams. 
Furthermore, LA-UAVs are capable of autonomously initiating their navigation systems to begin missions. This autonomous navigation integrates advanced obstacle avoidance and path planning algorithms, ensuring that the drones follow the shortest and safest routes during their operations.

\subsection{ Benefits of Proposed PDSR Architecture \label{sec:Benefits}} 

The major advantages of leveraging the proposed PDSR architecture can be summarized as follows: 
\begin{enumerate}
    \item Improving scalability and coverage, the proposed hierarchical (or multi-tiered) architecture enables scalable operations with HA-UAVs covering large areas at higher altitudes and LA-UAVs focusing on specific zones. This layered approach ensures efficient resource allocation, providing broad coverage while minimizing redundancy and overlap, unlike flat (or single-tier) architectures.  
    \item Supporting Swift and accurate detection for human under rubble. The proposed architecture is designed to optimize the speed and accuracy of detecting humans trapped under rubble. By enabling the rapid deployment of a network of UAVs, this system ensures efficient coverage of large areas through the use of HA-UAVs. Moreover, the integration of diverse sensor technologies on LA-UAVs allows these units to navigate closer to affected or high-risk buildings. This proximity, combined with the multi-sensor capabilities, significantly enhances the accuracy of human detection, making it possible to identify survivors in challenging environments.  
    \item Enabling effective multi-modal data analysis. The proposed architecture allows for implementing advanced Machine Learning (ML) models on both HA-UAVs and LA-UAVs. These models are specifically designed to handle multi-modal data, effectively filtering out noise, and focusing on meaningful information that indicate the presence of humans or other events of interest. This capability is crucial in disaster scenarios where data from multiple sources must be rapidly processed and interpreted to make timely decisions.  
    \item Allowing the UAVs to autonomously coordinate and navigate. By leveraging advanced path planning and obstacle avoidance algorithms, the proposed PDSR architecture empowers UAVs with the ability to autonomously coordinate and navigate through complex environments, while minimizing the time required to cover the target area, ensuring efficient and thorough coverage, and reducing the risk of collisions.   
    \item Establishing reliable communication with the DRC. Creating  dependable communication links between LA-UAVs, HA-UAVs, and the DRC is a cornerstone of the PDSR architecture. This ensures that critical data, such as the precise location of detected survivors and environmental hazards, is promptly forwarded to the DRC for further analysis and action. The reliability of this communication links is essential for coordinating rescue efforts and making informed decisions in real-time, thereby enhancing the overall effectiveness of the disaster response.  
\end{enumerate}

\section{PDSR Multi-modal Sensing  \label{sec:Solutions}} 

In this section, we discuss the primary design and implementation aspects of our developed framework. Then, we present  key results and provide a detailed discussion to evaluate the effectiveness and performance of the framework.

 \begin{figure*}[]
	\centering
		\scalebox{1.85}{\includegraphics[width=0.34 \textwidth]{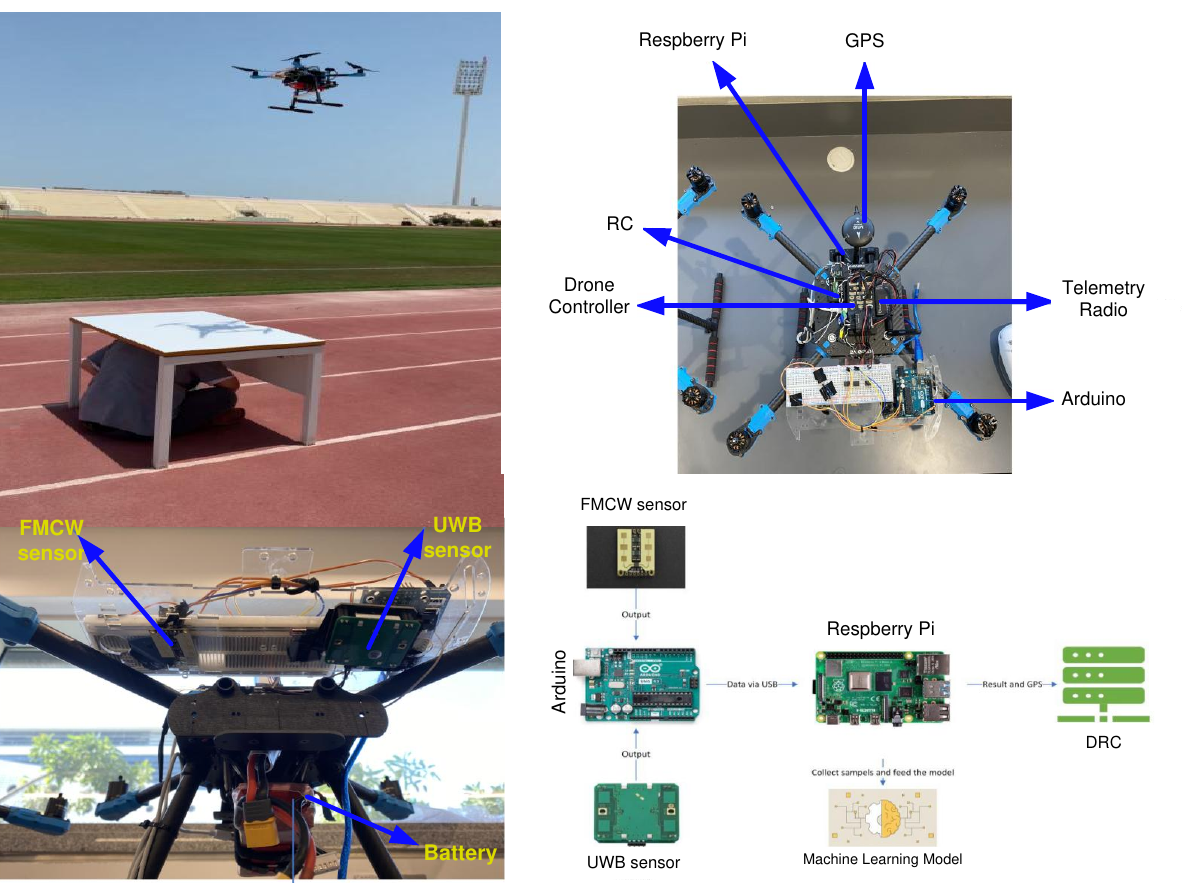}}
	\caption{Experiment setup 
 }
	\label{fig:Experiment}
\end{figure*}

\subsection{Design and Implementation}

Detecting humans beneath debris is challenging because the used sensors (or radar) must penetrate materials and be highly accurate to avoid false negatives. 
Thus, in our implementation, we explored a range of sensors based on prior research and existing literature, including cameras, radar sensors, PIR sensors, and voice sensors. The goal was to locate buried victims, extract relevant features, and perform classification to ensure high reliability and rapid response times. Initially, we found that cameras and thermal cameras are effective for detecting visible parts of trapped humans but cannot penetrate materials. These cameras require a direct line of sight, which may be obstructed in scenarios with obstacles or adverse environmental conditions.  
Voice recognition sensors require intensive signal processing and may not work if the survivor cannot scream or if environmental sounds are falsely detected. Therefore, we opted to explore two radar sensors: Frequency-Modulated Continuous Wave (FMCW) sensors and ultra-wideband (UWB) sensors. These radar sensors were chosen for their seamless integration into our system, as they can penetrate various materials, although the extent of penetration depends on the material properties, signal characteristics, and the altitude of the LA-UAVs used. 
Indeed, the FMCW sensor has the potential to detect buried survivors, while the UWB sensor can detect human presence through walls.  Moreover, these sensors are compact and lightweight, meeting the payload capacity limitations of the UAVs.


As shown in our experimental setup in Figure \ref{fig:Experiment},  a scaled down demonstration is implemented with the following parameters and constraints. The Holybro PX4 Development Kit – X500 v2, along with a 6500 mAh LiPo battery has been utilized. This kit features a quadcopter frame equipped with motors, propellers, ESCs, a power distribution board, an autopilot system, and a GPS module. For the radar sensors, we use the parallax x-band motion detector\footnote{https://www.parallax.com/product/x-band-motion-detector}, which leverages UWB technology, and the DfRobot mmWave 24GHz microwave radar distance sensor, which utilizes the FMCW principle. 
Furthermore, to perform edge computing tasks on the UAV, such as handling detection results, running navigation algorithms, and communicating with the DRC, we use a Raspberry Pi 4 Model B, which features a quad-core CPU, 8GB of RAM, and an integrated GPU. We also use an Arduino UNO microcontroller to serve as an interface between the computing unit and sensors. 
Finally, to establish 5G connectivity between the UAV and the DRC, a SIM card broadband device is attached to the UAV. This setup guarantees a 5G network connection for up to 10 hours before needing a recharge, ensuring sustained operation throughout the entirety of most missions.

In our experiments, it is assumed that the potential locations of the targets are identified by the HA-UAVs' cameras, as part of the damage assessment. Then, these locations are fed to the LA-UAVs, which use an autonomous navigation mechanism to reach the targets and start the sensing mission. For autonomous navigation, we have implemented an automated mission script to ensure the LA-UAVs navigate safely.  
During the sensing mission, the LA-UAV considers four main features: Doppler frequency, UWB detection output, FMCW readings, and UAV altitude. It is noteworthy that the UWB radar sensor can measure the Doppler frequency shift, which is the change in frequency of the reflected signal. This Doppler shift is then utilized to determine the UWB detection output. Specifically, a detection output of 1 indicates a human presence, while 0 indicates no detection, based on a predefined  threshold.   
The collected readings are sent to the computing unit via serial communication and processed by a multi-modal machine learning model to generate the final output, indicating whether a human is detected. This output, along with the GPS location, is then sent to the DRC to notify the SAR team on the ground if a human has been detected. 

\subsection{Results}

To assess the proposed framework, different experiments and scenarios were conducted at various UAV altitudes: 2m, 1.75m, and 1.5m. A total of 12,000 samples were collected, evenly divided between detection and non-detection scenarios, with 2,000 samples per altitude. In the first scenario, we assumed perfect drone stability and unaffected sensor readings. In the second scenario, we maintained stability but added bricks as rubble to test the model's performance with different materials. In the third scenario, we collected data while the UAV was hovering to assess accuracy degradation due to instability, resulting in a final dataset combining stable and hovering data. 
Furthermore, three ML models—logistic regression (LR), random forest (RF), and decision tree (DT)—were used for human detection. These models balance good performance with the processing capacity of the hosting computer. To enhance model recall, multiple output samples from these machine learning models were considered to calculate the probability of detection. 

First, we evaluate the performance of our framework in detecting humans beneath debris, using various ML models and data collected from the first two scenarios. As shown in Table \ref{tab:ex1}, leveraging multi-modal data enables different ML models to achieve good detection accuracy. Specifically, the RF classifier demonstrates the highest accuracy and recall. Recall is particularly important for our application, as it prioritizes the model's sensitivity to detecting humans, even at the expense of increasing false positives. Therefore, we have selected the RF model for our detection tasks. However, our framework is flexible and can easily incorporate different ML/deep learning models as needed.

\begin{table}[b!]
	\centering
\caption{The obtained detection accuracy for different ML models.  }
	  \label{tab:ex1}
\begin{tabular}{|c|c|c|c|c|c|c|c|c|c|} 
			\hline 
\textbf{ML Model} & \textbf{Accuracy} & \textbf{Precision} & \textbf{Recall} & \textbf{F1 Score}  
\\
\hline
\textbf{LR } & 97.07 \% & 97.73 \% & 98.22 \% & 98.70 \%
  \\
\hline
\textbf{DT} & 98.69 \% & 98.50 \% & 98.89 \% & 98.70 \%  
  \\ 
\hline
\bf{RF } & 98.75 \% & 98.40 \% & 99.11 \% & 98.70 \%   
  \\   
\hline
\end{tabular} 
\end{table}

\begin{figure}[t!]
	\centering
		\scalebox{1.32}{\includegraphics[width=0.33 \textwidth]{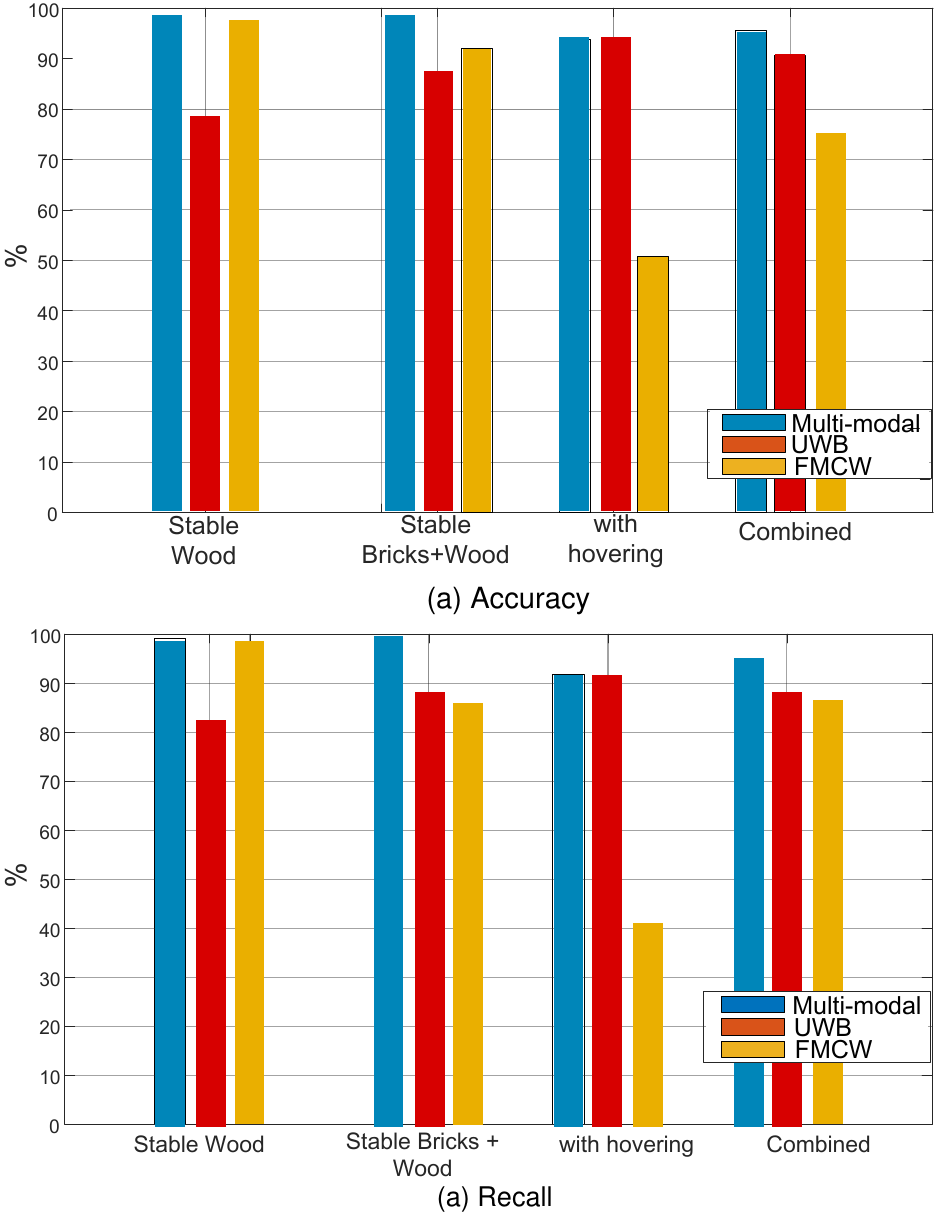}}
	\caption{The obtained accuracy and Recall for different considered scenarios.  }
	\label{fig:table20}
\end{figure}

Second, we evaluated the detection accuracy and recall of our framework across various scenarios and compared them to single-modality approaches, as illustrated in Figure \ref{fig:table20}.  These scenarios include: Stable Wood, Stable Wood and Bricks, Wood and Bricks with Hovering, and a Combined Dataset.  
In the first two scenarios, we assumed perfect drone stability, with sensors collecting data while being stably hung. The scenarios involved different types of rubble—wood and a combination of wood and bricks, respectively—to assess how the model performs across various materials.   
In the third scenario, we examined the impact of drone movement by collecting data while the drone was hovering. This allowed us to observe how sensor instability affects detection accuracy, while increasing the complexity of the environment. Finally, all collected data from these scenarios were aggregated into a combined dataset to evaluate the overall robustness and performance of the model under diverse conditions.  
As shown in Figure \ref{fig:table20}, our solution  consistently performs well across different materials and conditions, significantly outperforming single-modality schemes that rely solely on UWB output or FMCW readings.  Notably, our solution, when trained on the combined dataset, achieves  a robust accuracy of 94.1\%, despite experiencing a slight drop in accuracy during drone hovering. This decline is expected due to false positive readings caused by sensor motion. Moreover, it is evident that recall is more significantly affected by hovering than detection accuracy, underscoring the challenge of maintaining sensitivity with less stable drones.   

\begin{figure}[t!]
	\centering
		\scalebox{1.16}{\includegraphics[width=0.335 \textwidth]{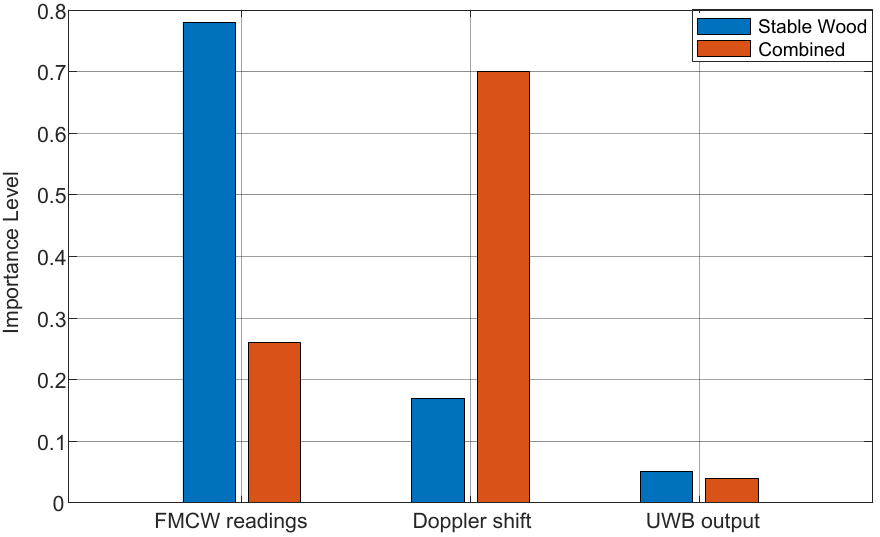}}
	\caption{Collected features' importance on different scenarios.  }
	\label{fig:fig30}
\end{figure}

Finally, we evaluate the significance of each feature in contributing to the overall detection accuracy in Figure \ref{fig:fig30}. This analysis is crucial when working with multi-modal data, as it allows us to determine the relative importance of different modalities. By understanding which features most significantly impact the model's performance, we can optimize the learning process. This involves prioritizing the more influential modalities that enhance accuracy, while potentially disregarding those that contribute less or may introduce noise. 
Interestingly, Figure \ref{fig:fig30} depicts that in stable environment, FMCW readings are the most significant features in contributing to the overall detection accuracy. On the contrary, in unstable environments, Doppler shift generated by the UWB sensor plays a significant role in obtaining the final accuracy.   

\subsection{Discussion}

The proposed PDSR architecture enables UAVs equipped with diverse cameras and sensor technologies to efficiently conduct continuous survivor detection for both visible and non-visible targets buried under debris. Specifically, we envision detection being executed in two phases: visible target detection and non-visible target detection. 
In the first phase, HA-UAVs utilize high-resolution cameras to identify visible humans or targets.  In the second phase, LA-UAVs equipped with radar systems detect the presence of non-visible individuals.  
In this context, HA-UAVs offer several key advantages: they have lower latency, provide superior image resolution due to their closer proximity to the ground (compared to balloons or satellites), and can easily return to base for maintenance or payload adjustments. Furthermore, HA-UAVs are known for their rapid deployment, surpassing other platforms like balloons, airships, and satellites in responsiveness.   

At LA-UAVs, leveraging both FMCW and UWB radars depicts significant potential in enhancing detection accuracy while reducing response time. FMCW radar continuously monitors variations in the phase and frequency of reflected signals, enabling the detection of micromotions such as human vital signs. However, FMCW radar requires stable mounting due to its high sensitivity to slight movements, making it more challenging to use compared to other radar types. 
On the other hand, UWB radar emits a short-duration impulse signal that spans a broad frequency spectrum, then awaits the return signal reflected from the target object. This received signal contains critical information about the object, particularly in identifying human movement, including heartbeat and respiration.

By applying various signal processing and machine learning algorithms to both FMCW and UWB readings, we were able to detect human presence with high accuracy. One notable advantage of UWB radar is its operation around the 10 GHz frequency, which provides strong penetration capabilities with minimal power loss during signal transmission. Furthermore, UWB radar demonstrated better resilience to vibrations, though it requires more pronounced movements to effectively detect motion or presence. 
Indeed, both FMCW and UWB radars can generally penetrate wooden materials without significant performance loss. However, they encounter difficulties when dealing with concrete and different types of bricks, especially at thicknesses of 15 cm or more.   

Finally, the spatial distribution of damaged areas and victims significantly affects the type and number of UAVs required for post-disaster operations. For example, if victims are concentrated in specific areas, deploying a single HA-UAV per cluster along with a few LA-UAVs may be sufficient. However, in situations where affected areas are widely dispersed, a greater number of both HA and LA-UAVs will be necessary to expedite PDSR operations. Therefore, anticipating critical areas of interest is essential. This involves leveraging predictive analytics and coordinating various technologies to effectively detect, identify, and address these areas.

\section{Conclusions and Future Directions\label{sec:conclusion}}

In this paper, we present a novel PDSR system architecture that enables effective, swift, and large-scale detection and rescue of disaster victims. By leveraging various types of UAVs, our system overcomes PDSR challenges such as damaged infrastructure, unstable ground, and blocked roads. UAVs can rapidly access areas inaccessible to ground vehicles, facilitating faster detection of survivors and efficient data collection for rescue teams. This technology reduces the need for human intervention in SAR operations, minimizing the latency and risks faced by rescue teams.
Specifically, adopting the proposed multi-tier UAV swarm architecture significantly enhances victim detection under rubble, thereby increasing survival rates. Integrating multiple sensing technologies significantly enhances detection accuracy, ensuring precise identification of survivors. For instance, the proposed detection model achieves an accuracy rate of 94\%.

The presented work opens up several promising research directions for future exploration, which call for extensive experiments to evaluate and utilize different sensor technologies, optimize the number and placement of UAVs, and implement efficient algorithms for multi-modal data analysis. These future directions include:
{   
\subsubsection{ Levraging Terahertz (THz) imaging technology } 
Terahertz (THz) imaging holds great potential as a promising technology for PDSR missions, particularly in scenarios where conventional methods fall short. Thus, we envision that developing a novel design that ingeniously uses a communication waveform for sensing within the THz frequency band is worth further investigation. This innovative approach can seamlessly combine communication and sensing functionalities, leveraging the uplink communication waveform for opportunistic sensing. Through this methodology, crucial environmental sensing parameters can be directly extracted from received signal strength using techniques like tensor theory \cite{chaccour2024joint}. 
Furthermore, non-terrestrial infrastructures enable the flexible deployment of various THz sensing equipment at optimal altitudes and locations. However, the practical deployment of THz non-terrestrial networks (NTN) faces several technical challenges, including but not limited to the feasibility assessment of THz frequencies for space-to-earth links and transceiver implementation \cite{jiang2024terahertz}. 
  

While Terahertz (THz) waves hold promising potential for penetrating certain non-conducting materials and sensing beyond obstacles, the current research on combined sensing and communication systems in the THz frequency range lacks a comprehensive methodology to manage uncertainties within THz channels. This highlights the necessity for an integrated approach that leverages AI alongside sensing techniques to address this challenge effectively. 

\subsubsection{Implementing efficient algorithms for multi-modal data analysis} 
Leveraging multi-modal data, such as combining RF signals from mobile phones \cite{depold2023direction} with sensor data like sound, motion, radar, acoustic, seismic, and magnetic signals, can greatly enhance the accuracy and robustness of PDSR operations. This integration provides emergency responders with comprehensive insights, enabling more effective and timely rescue efforts. 
For instance, integrating RF data with radar, acoustic, and magnetic sensor information improves the detection of buried victims and distinguishes between survivors and casualties, even when RF signals are weak or obstructed. This approach helps rescuers better understand disaster zones and prioritize rescue operations. 
In this context, developing innovative algorithms and advanced machine learning techniques to fuse and analyze multi-modal data is crucial. This enables the creation of predictive models that anticipate hazards and optimize response strategies in real-time.  
This interdisciplinary approach, combining signal processing, machine learning, and disaster management, is a crucial step forward in improving global disaster resilience and response effectiveness.
} 

\subsubsection{Optimizing UAV deployment for enhanced cooperative sensing} 
Cooperative multi-UAV sensing has the potential to allow post-disaster management teams to enhance their capabilities, mitigate disaster impacts, and save lives. This can be achieved by leveraging multiple UAVs simultaneously to perform radar detection from different locations and angles to sense the target area. 
Operating cooperatively, the UAVs can be viewed as a distributed MIMO radar system, enabling precise localization of objects in the target area. However, this approach raises several important questions: How many LA-UAVs are needed to minimize scanning and detection time while maximizing detection accuracy? How can we effectively coordinate between HA-UAVs and LA-UAVs? 
Addressing these questions necessitates the development and implementation of efficient algorithms to optimize the number and locations of HA-UAVs and LA-UAVs, thereby maximizing coverage and detection accuracy.

\balance 

\bibliographystyle{IEEEtrannames}
\bibliography{OCL}

\end{document}